\documentclass[10pt,twocolumn,letterpaper]{article}

\usepackage[pagenumbers]{iccv} % To force page numbers, e.g. for an arXiv version

\definecolor{iccvblue}{rgb}{0.21,0.49,0.74}
\usepackage[pagebackref,breaklinks,colorlinks,allcolors=iccvblue]{hyperref}

\usepackage{multirow}
\usepackage{bm}

\DeclareMathOperator*{\argmin}{arg\,min}
\usepackage{graphicx}
\usepackage[normalem]{ulem}
\usepackage{pifont}
\newcommand{\cmark}{\ding{51}}%
\newcommand{\xmark}{\ding{55}}
\usepackage{colortbl}

\usepackage{amsfonts}
\usepackage{algorithmic}
\usepackage{algorithm}
\usepackage{array}
\usepackage{textcomp}
\usepackage{stfloats}
\usepackage{url}
\usepackage{verbatim}
\usepackage{color}
\usepackage{multirow}
\usepackage{bbm}
\usepackage{bm}
\usepackage{makecell}

\usepackage{wrapfig}
\usepackage{marvosym}

\title{Bi-Level Optimization for Self-Supervised AI-Generated Face Detection}

\author{
Mian Zou$^{1,2}$, \
Nan Zhong$^{2}$, \
Baosheng Yu$^{3}$, 
Yibing Zhan$^{4}$, and
Kede Ma$^{2,}$\thanks{Corresponding author.} \\
$^1$Jiangxi University of Finance and Economics \
$^2$City University of Hong Kong \\
$^3$Nanyang Technological University \
$^4$Yunnan United Vision Technology
\\
{\tt\small mianzou2-c@my.cityu.edu.hk, \{nzhong, kede.ma\}@cityu.edu.hk,} \\
{\tt\small baosheng.yu@ntu.edu.sg, zybjy@mail.ustc.edu.cn}\\
\tt\small \url{https://github.com/MZMMSEC/AIGFD_BLO}
}

\begin{document}
\maketitle
\begin{abstract}
AI-generated face detectors trained via supervised learning typically rely on synthesized images from specific generators, limiting their generalization to emerging generative techniques. To overcome this limitation, we introduce a self-supervised method based on bi-level optimization. In the inner loop, we pretrain a vision encoder only on photographic face images using a set of linearly weighted pretext tasks: classification of categorical exchangeable image file format (EXIF) tags, ranking of ordinal EXIF tags, and detection of artificial face manipulations. The outer loop then optimizes the relative weights of these pretext tasks to enhance the coarse-grained detection of manipulated faces, serving as a proxy task for identifying AI-generated faces. In doing so, it aligns self-supervised learning more closely with the ultimate goal of AI-generated face detection. Once pretrained, the encoder remains fixed, and AI-generated faces are detected either as anomalies under a Gaussian mixture model fitted to photographic face features or by a lightweight two-layer perceptron serving as a binary classifier. Extensive experiments demonstrate that our detectors significantly outperform existing approaches in both one-class and binary classification settings, exhibiting strong generalization to unseen generators.
\end{abstract}

\section{Introduction}
\label{sec:intro}
The synthesis of human faces by modern generative models---ranging from generative adversarial networks (GANs)~\cite{karras2020analyzing, esser2021taming} to diffusion models~\cite{rombach2022high, song2021denoising, yu2023freedom, wu2023better, Midjourney, podell2024sdxl}---has reached a level of realism that renders AI-generated faces virtually indistinguishable from ``real'' face photographs~\cite{nightingale2022ai}. While this progress unlocks enormous creative and practical potential, it also amplifies risks related to misinformation, privacy breaches, and cybercrime. As new generators emerge, the urgent need for robust detectors of AI-generated faces has never been greater.

Most existing detectors rely on supervised learning, treating the problem as end-to-end feature learning and binary classification between photographic and AI-generated\footnote{In this paper, we refrain from using the terms ``real'' and ``fake,'' since it is challenging to operationalize a precise definition of ``a real face'' in the context of image generation~\cite{zou2024semantic}.} faces~\cite{liu2020global, wang2023dire, wang2019cnngenerated, liu2024forgeryaware, tan2023learning,durall2020watch, corvi2023intriguing, dong2022think, frank2020leveraging}. Unfortunately, these methods latch onto artifacts specific to the generators seen during training, and thus struggle to adapt to novel generative techniques. An alternative line of work seeks model-agnostic cues, \eg, physiological inconsistencies in head poses~\cite{yang2019exposing}, pupil shapes~\cite{guo2022icassp_eyes}, or corneal highlights~\cite{hu2021exposing}. However, handcrafted features may miss the subtle statistical differences introduced by state-of-the-art generators. More recently, self-supervised features learned solely from photographic images~\cite{ojha2023towards,cozzolino2024zed} have demonstrated better cross-generator performance. Yet, because their pretext tasks are not explicitly tailored to AI-generated face detection, these approaches can still fall short of optimal.

To bridge this gap, we build on the prior work~\cite{zou2025self} by introducing a bi‑level optimization scheme~\cite{dempe2002foundations} that explicitly aligns self‑supervised pretraining with AI‑generated face detection. In the inner loop, we train a vision encoder on face photographs using a linearly weighted combination of pretext tasks, including 1) classification of \textit{categorical} exchangeable image file format (EXIF) tags (\eg, \texttt{white balance mode} and \texttt{flash}), 2) ranking of \textit{ordinal}  EXIF tags (\eg, \texttt{ISO speed} and \texttt{aperture}), and 3) detection of artificial face manipulations (\eg, flipped or affine-warped regions). In the outer loop, we dynamically adjust the relative task weights to refine the feature space for coarse-grained \textit{manipulated} face detection. This proxy task enables us to better identify \textit{AI-generated} faces without ever incorporating them during self-supervision.

Once pretraining is complete, the encoder is frozen. AI-generated faces can then be uncovered either by treating them as anomalies under a Gaussian mixture model (GMM) fitted to photographic face features, or by training a lightweight two-layer perceptron on the learned representations. Through extensive experiments in both one-class and binary classification settings, our Bi-Level AI-generated face DEtector with Self-supervision (BLADES) significantly outperforms prior approaches and generalizes to unseen generators.

In summary, our contributions are threefold.
\begin{itemize}
    \item We present BLADES, a bi-level optimization scheme that explicitly steers self-supervised pretraining toward AI-generated face detection.
    \item We implement BLADES using joint embedding~\cite{zou2024sjedd} that incorporates EXIF-based and manipulation-based pretext/surrogate tasks to detect AI-generated faces.
    \item We demonstrate state-of-the-art performance in both one-class and binary classification evaluations, with strong cross-generator generalization.
\end{itemize}

\section{Related Work}
In this section, we provide a concise overview of AI-generated face detection, self-supervised learning, and EXIF metadata for computer vision.

\subsection{AI-Generated Face Detection}
\noindent\textbf{Model-Dependent Detectors} exploit characteristic artifacts introduced by specific generators, \eg, spatial textures inconsistencies~\cite{liu2020global}, frequency distortions from upsampling~\cite{corvi2023intriguing, durall2020watch, dong2022think, frank2020leveraging, liu2024forgeryaware}, or reconstruction errors in diffusion models~\cite{wang2023dire,luo2024lare}. While highly effective against known generative methods, these approaches tend to overfit to the model peculiarities they have been trained on, and thus struggle to generalize as new generators emerge.

\noindent\textbf{Model-Agnostic Detectors} seek cues that transfer across generators. 
Early work framed AI-generated image detection as a one-class classification (also known as anomaly detection) problem~\cite{hawkins1980identification}, leveraging physiological markers such as implausible head poses~\cite{yang2019exposing}, pupil shape anomalies~\cite{guo2022icassp_eyes}, and corneal specular highlights~\cite{hu2021exposing}. 
More recently, self-supervised representations learned from purely photographic images---via pretext tasks like image-text alignment~\cite{ojha2023towards}, denoising~\cite{liu2022detecting}, super-resolution~\cite{cozzolino2024zed}, and EXIF-based ordinal ranking~\cite{zou2025self}---have shown stronger cross-generator performance.
However, because these pretext tasks are designed without explicit regard for the end goal of AI-generated face detection, they are bound to be suboptimal. Here, we present BLADES that leverages bi‑level optimization to tailor pretext tasks to the detection objective, relying exclusively on face photographs.

\begin{figure*}[]
  \centering
  \includegraphics[width=1\linewidth]{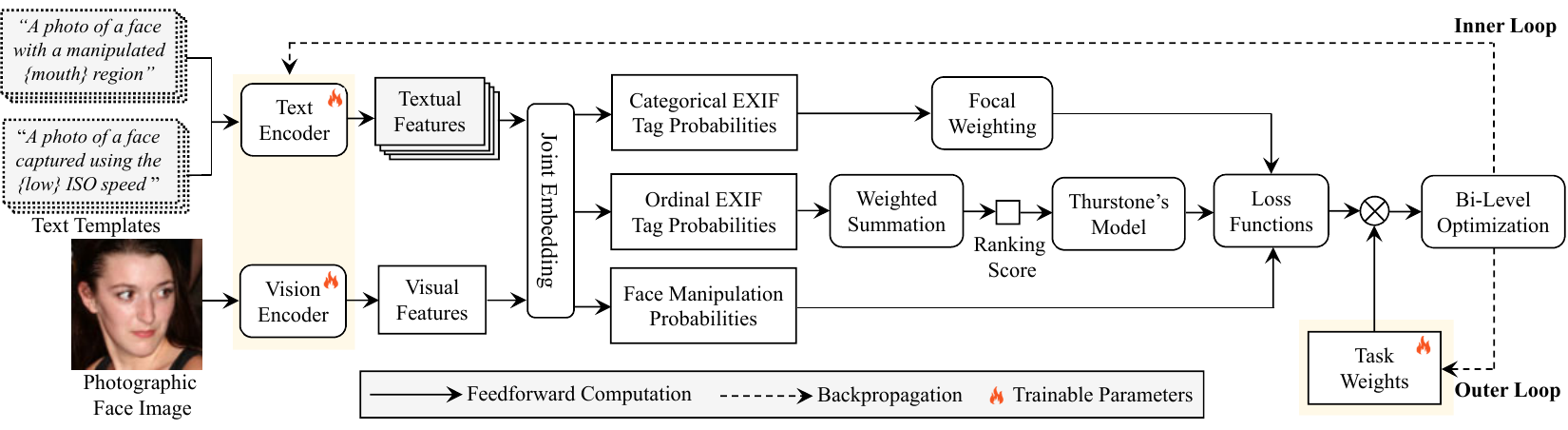}
  \caption{System diagram of the proposed BLADES for AI-generated face detection.}
  \label{fig: SSL}
\end{figure*}

\subsection{Self-Supervised Learning}
Self-supervised learning has proven a powerful paradigm for acquiring transferable visual features by solving surrogate tasks, \eg, rotation prediction~\cite{gidaris2018unsupervised}, jigsaw puzzles~\cite{goyal2019scaling}, contrastive learning~\cite{radford2021CLIP}, and masked image modeling~\cite{he2021masked}. Such methods have driven advances across
 face parsing~\cite{zheng2022general}, object segmentation~\cite{ziegler2022self} and detection~\cite{yang2021instance}, and forensic analysis~\cite{zheng2023exif}. Yet, a critical gap remains: conventional self-supervised learning pipelines do not explicitly align their pretext losses with downstream tasks, leading to suboptimal specialization.

To address this, recent work has explored task-aware pretext design~\cite{huh2018fighting}, fine-tuning initialization~\cite{wu2022noisytune}, modular pretraining~\cite{li2023aligndet}, and progressive domain adaptation~\cite{reed2022self}. However, multi‑stage training schemes often depend on ad hoc heuristics. By contrast, bi-level optimization~\cite{zakarias2024bissl, chen2023structure, somayajula2023bi}, as employed in this paper, provides a principled and mathematically grounded approach for steering self-supervised pretraining end-to-end toward the downstream objective.

\subsection{EXIF Metadata for Computer Vision}
A promising yet under-explored source of supervisory signal is EXIF metadata---the standardized camera tags embedded in image files. Originally used for camera calibration~\cite{zhang2000flexible}, EXIF information has since supported image quality assessment~\cite{fang2020perceptual} and image splicing detection~\cite{huh2018fighting}. Recent work has shown that encoding EXIF metadata as a linguistic modality and jointly embedding it with images yields representations that transfer to forgery detection~\cite{zheng2023exif}. Here, we extend the prior EXIF-based learning~\cite{zou2025self} by incorporating a richer set of categorical and ordinal tags, integrating face manipulation tasks, and employing bi-level optimization to align these pretext objectives with a surrogate task of AI-generated face detection.

\section{Proposed Method: BLADES}
This section presents in detail the proposed BLADES, whose system diagram is shown in Fig.~\ref{fig: SSL}.

\subsection{Problem Formulation}
We seek to learn a vision encoder that, without ever seeing synthesized faces, produces representations tailored for AI-generated face detection.

Given a training minibatch of face photographs $\mathcal{B}_\mathrm{tr}$ for self-supervision and a separate validation minibatch $\mathcal{B}_\mathrm{val}$, we formulate a bi-level optimization problem:
\begin{subequations}\label{eq:auto-weighting}
\begin{align}\label{eq:primary}
& \min_{\bm \lambda} \sum_{\bm x\in \mathcal{B}_{\mathrm{val}}} \ell_1 \left (\bm  x; {\bm \theta}^\star \right) \\
\label{eq: all_tasks}
\text{s.t.}\quad {\bm \theta}^\star &= \argmin_{\bm \theta}  \sum_{\bm x\in \mathcal{B}_{\mathrm{tr}}}  \sum_{i=1}^{K} \lambda_i \ell_i(\bm x; \bm \theta).
\end{align}
\end{subequations}
Here $\ell_1$ denotes the loss for a surrogate primary task for AI-generated face detection and $\ell_i$, for $i=2,\ldots, K$, represents the self-supervised pretext losses. 
The vector ${\bm\lambda}=[\lambda_1, \ldots, \lambda_K]$ weights each task, while $\bm\theta$ comprises all learnable parameters. By alternating updating $\bm\theta$ (in the inner loop) and $\bm\lambda$ (in the outer loop), the optimization automatically prioritizes those pretext objectives most beneficial to downstream detection.

\subsection{Task Definition  and Implementation}~\label{subsec: pretext_tasks}
To endow our vision encoder with the ability to distinguish AI-generated faces without direct exposure, we define a suite of self-supervised tasks---integrated via joint image-text embedding~\cite{radford2021CLIP}---that capture both camera-specific characteristics and face-centric alterations. Concretely, we organize our tasks into four sets: 1) a surrogate primary task that approximates AI-generated face detection through coarse-grained face manipulation detection, 2) categorical EXIF tag classification~\cite{huh2018fighting}, 3) ordinal EXIF tag ranking~\cite{zou2025self}, and 4) fine-grained face manipulation detection~\cite{bohavcek2023geometric}. Each introduces a dedicated loss, and their weighted sum drives the inner-loop pretraining in Eq.~\eqref{eq: all_tasks}.

\noindent\textbf{Joint Embedding.} We embed each face image $\bm x$ and text template $\bm t$ into a shared $\mathbb{R}^N$ space~\cite{zou2024sjedd} using a vision encoder $\bm f_{\bm\phi}$ and a text encoder $\bm f_{\bm \varphi}$, computing
\begin{align}
    \label{eq:unnorm_similarity}
    s(\bm x, \bm{t})= \frac{\langle\bm f_{\bm \phi}(\bm x),\bm f_{\bm \varphi}(\bm{t})\rangle}{\tau},
\end{align}
where $\langle\cdot,\cdot\rangle$ represents the inner product, and $\tau$ is a learnable temperature. All tasks leverage these similarity scores to produce probability estimates for their respective labels.

\begin{figure}[]
  \centering
  \subfloat[]{\includegraphics[width=0.2\linewidth]{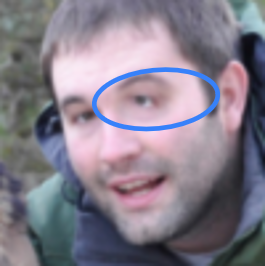}}
  \subfloat[]{\includegraphics[width=0.2\linewidth]{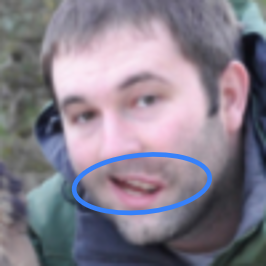}}
  \subfloat[]{\includegraphics[width=0.2\linewidth]{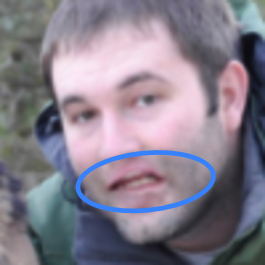}}
  \subfloat[]{\includegraphics[width=0.2\linewidth]{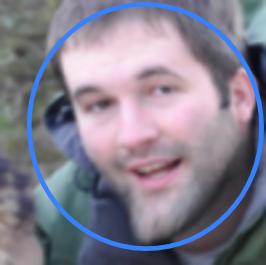}}
  \subfloat[]{\includegraphics[width=0.2\linewidth]{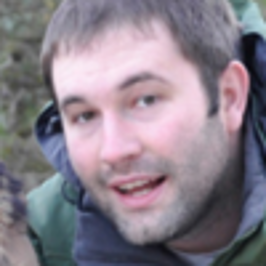}}
  \caption{Visualization of artificially manipulated face photographs by \textbf{(a)} horizontal eye flipping, \textbf{(b)} horizontal mouth flipping, \textbf{(c)} vertical mouth flipping, and \textbf{(d)} global affine transformation, respectively. \textbf{(e)} The original face is also shown as reference.}
  \label{fig: face_manipulations}
\end{figure}

\noindent\textbf{Task 1: Coarse-Grained Face Manipulation Detection.}
As a proxy for AI-generated face detection, we generate manipulated faces via local flips and global affine transforms~\cite{zou2025self} (see Fig.~\ref{fig: face_manipulations}). The  text template is defined as $\bm t_g$: ``\textit{A photo of a \{$g$\} face},'' where $g \in \mathcal{G} = \{\texttt{manipulated}, \texttt{photographic}\}$. We apply a fidelity loss~\cite{tsai2007frank, zou2024sjedd}:
\begin{equation}
\label{eq:cls_face_binary}
\ell_1(\bm{x}) = 1 - \sum_{g\in\mathcal{G}}\sqrt{p(g\vert\bm{x})\hat{p}(g\vert\bm{x})},
\end{equation}
where $p(g\vert\bm{x})=1$ if $\bm x$ is a manipulated face and zero otherwise, and the model's predicted probability $\hat{p}(g\vert\bm{x})$ is obtained by applying a softmax over the similarity scores:
\begin{equation}\label{eq: softmax_prob}
    \hat{p}(g\vert\bm x) = \frac{\exp\left(s(\bm x, \bm t_g)\right)}{\sum_{g' \in \mathcal{G}} \exp\left(s(\bm x, \bm t_{g'})\right)}.
\end{equation} 

\noindent\textbf{Task 2: Categorical EXIF Tag Classification.} 
To equip our encoder with camera-specific priors, we treat the prediction of each selected categorical EXIF tag (\eg, \texttt{makes}) as a self-supervised classification task. Given the $i$-th categorical tag with label set $\mathcal{C}_i$, we structure the text template $\bm t_{i,c}$ as \textit{``A photo of a face captured using the \{c\} \{i-th categorical tag\},''} where $c\in \mathcal{C}_i$, and embed both the image $\bm x$ and $\bm t_{i,c}$ into our shared vision-language space.
We then convert the resulting similarity scores into a class probability distribution via softmax (same way as Eq.~\eqref{eq: softmax_prob}). This yields $\hat{p}_i(c\vert\bm x)$, the model's predicted probability that the $i$-th EXIF tag of image $\bm x$ was set to the value $c$.

In practice, EXIF tag values exhibit long-tailed distributions (\eg, \texttt{Canon} images may outnumber \texttt{Apple} by $10\times$), which can bias training toward majority classes. Thus, we employ a focal version~\cite{lin2017focal} of the multi-class fidelity loss that down-weights well-classified samples:
\begin{align}\label{eq:focal_cls_exif}
 \ell_i(\bm{x})
= \left(1-\hat{p}_i(y\vert\bm{x})\right)^\gamma \left(1 - \sum_{c\in\mathcal{C}_i} \sqrt{ p_i(c\vert\bm{x})\hat{p}_i(c|\bm{x}) }\right), 
\end{align}
where $y$ is the ground-truth label, and $p_i(c\vert\bm x) = \mathbb{I}\{c=y\}$. 
% Finally, summing over all categorical tags gives the total Task 2 loss.

\noindent\textbf{Task 3: Ordinal EXIF Tag Ranking.} 
Because raw numerical EXIF values (\eg, \texttt{ISO 100} vs. \texttt{ISO 200}) may be poorly handled by text tokenizers---which tend to fragment or ignore numeric relationships---we first represent the $j$-th ordinal tag using one of three textual levels $\mathcal{L} = \{\texttt{low}, \texttt{medium},\texttt{high}\}$. This gives rise to the text template $\bm t_l$: ``\textit{A photo of a face captured using the \{$l$\} \{$j$-th ordinal tag\},}'' where $l\in\mathcal{L}$. Each image $\bm x$ is paired with these level-specific text templates and embedded into the joint feature space. To recover a continuous estimate of the original EXIF value, we compute a probability-weighted sum over these discretized levels: 
\begin{align}\label{eq: weightes_sum}
    \hat{r}_j(\bm{x}) = \sum_{l\in\mathcal{L}} \hat{p}_j(l\vert\bm x) \times d(l),
\end{align}
where $\hat{p}_j(l\vert\bm x)$ is again computed from the similarity scores in Eq.~\eqref{eq:unnorm_similarity} via softmax, and 
\begin{align}
    d(l) = 
\begin{cases}
1 & \text{if } l = \texttt{low} \\
2 & \text{if } l = \texttt{medium}\\
3 & \text{if } l = \texttt{high}
\end{cases}
\end{align}
is the corresponding numerical level. This preserves fine-grained ordinal information while 
leveraging the strengths of our joint embedding architecture.

To enforce the correct ordering of EXIF tags, we perform pairwise ranking. Given two images $(\bm x, \bm x')$, the ground-truth relation is $p_j(\bm{x}, \bm{x}') = 1$ if $\mathrm{tag}_j(\bm{x})\ge \mathrm{tag}_j(\bm{x}')$ and zero otherwise. Under the Thurstone's model~\cite{thurstone1927law}, the predicted probability that the value of the $j$-th EXIF tag of image $\bm{x}$ is greater than that of $\bm{x}'$ is 
\begin{align}\label{eq:thurstone}
\hat{p}_j(\bm{x}, \bm{x}')= \Phi\left(\frac{\hat{r}_j(\bm{x}) - \hat{r}_j(\bm{x}')}{\sqrt{2}}\right),
\end{align}
where $\Phi(\cdot)$ is the standard normal cumulative distribution function. Similarly, we minimize a fidelity-based ranking loss for each tag:
\begin{align}\label{eq: fidelity}
\ell_j(\bm{x}, \bm{x}') =& 1 - \sqrt{p_j(\bm{x}, \bm{x}')\hat{p}_j(\bm{x}, \bm{x}')} \nonumber \\
&- \sqrt{\left(1-p_j(\bm{x}, \bm{x}')\right)\left(1-\hat{p}_j(\bm{x}, \bm{x}')\right)}.
\end{align}
% Aggregating over all ordinal tags with their learned weights $\{\lambda_j\}$ aligns the embedding space with the intrinsic ordering of camera settings.

\noindent\textbf{Task 4: Fine-Grained Face Manipulation Detection.} To improve the encoder's sensitivity to localized tampering, we introduce a fine-grained classification task that identifies which face regions have been manipulated. Let $\mathcal{M}=\{\texttt{eye}, \texttt{mouth}, \texttt{nose}\}$ be the set of target regions. For each $m\in\mathcal{M}$, we define the text template $\bm t_m$: \textit{``A photo of a face with a manipulated \{$m$\} region,''} and compute the similarity score $s(\bm x, \bm t_m)$. These scores are converted into region-specific manipulation probabilities $\hat{p}(m\vert\bm{x})=\mathtt{sigmoid}\left(s(\bm{x}, \bm{t}_m)\right)$. Denoting the ground-truth indicator $p(m\vert\bm x) = 1$ if region $m$ of $\bm x$ is manipulated and zero otherwise, we then measure a fidelity loss for each region:
\begin{align}\label{eq: cls_face_multilabel}
\ell_m(\bm{x})
=&1 - \sqrt{p(m\vert\bm{x})\hat{p}(m\vert\bm{x})} \nonumber \\ 
&-\sqrt{\left(1-p(m\vert\bm{x})\right)\left(1 - \hat{p}(m\vert\bm{x})\right)}.
\end{align}
% Averaging over all regions yields the fine-grained manipulation loss. 
By including coarse- and fine-grained pretext tasks, the bi-level optimization scheme is guided to learn a photographic face feature space that aligns AI-generated face detection with EXIF-based supervision.
% both global and local manipulation cues in service of robust AI-generated face detection.

\subsection{Bi-Level Task Alignment}
To explicitly steer self-supervised pretraining toward AI-generated face detection, we employ a bi-level optimization strategy that alternately updates the encoder parameters  $\bm{\theta} = \{\bm{\phi}, \bm{\varphi}, \tau\}$ and task weights $\bm{\lambda}$ in each iteration~\cite{liu2022auto_lambda}.

\noindent\textbf{Inner-Loop Update (Feature Learning).} We update the encoder parameters $\bm \theta$ by minimizing the weighted sum of all pretext losses on a training minibatch $\mathcal{B}_\mathrm{tr}$:
\begin{align}
            \bm \theta' = \bm \theta - \alpha \nabla_{\bm \theta}\sum_{\bm x\in \mathcal{B}_{\mathrm{tr}}}\sum_{i=1}^K \lambda_i \ell_i(\bm x; \bm \theta), \label{subeq:theta'} 
\end{align}
where $\alpha$ is the inner-loop learning rate.

\noindent\textbf{Outer-Loop Update (Task Weighting).} We adjust the task weight vector $\bm \lambda$ to minimize the surrogate primary loss $\ell_1$ on a validation minibatch  $\mathcal{B}_\mathrm{val}$:
\begin{align}
     \bm \lambda \leftarrow \bm \lambda - \beta \nabla_{\bm \lambda}\sum_{\bm x\in \mathcal{B}_{\mathrm{val}}}\ell_1(\bm x; \bm \theta'(\bm \lambda)), \label{subeq:lambda_update} 
\end{align}
where $\beta$ is the outer-loop learning rate, and we make explicit the dependence of $\bm\theta'$ on $\bm \lambda$. This step dynamically emphasizes those pretext objectives that most improve the surrogate detection task.

\noindent\textbf{Inner-Loop Refinement.} Finally, we re-optimize $\bm \theta$ using the updated $\bm \lambda$, further aligning the learned representations with the detection objective:
\begin{align}
        \bm \theta \leftarrow \bm \theta - \alpha \nabla_{\bm \theta}\sum_{\bm x\in \mathcal{B}_{\mathrm{tr}}}\sum_{i=1}^K \lambda_i \ell_i(\bm x; \bm \theta)\label{subeq:theta_update}.
\end{align}

\begin{table*}[]
\centering
\small
\resizebox{0.9\linewidth}{!}{
\begin{tabular}{lcccccccccc}
\toprule
Method & StyleGAN2 & VQGAN & LDM & DDIM & SDv2.1 & FreeDoM & HPS & Midjourney & SDXL & Average \\
\hline
CNND~\cite{wang2019cnngenerated} & 50.61 & \cellcolor{lightgray!20}{99.89} & 53.07  & 56.55 & 50.51  & 58.62  & 50.31  & 51.66  & 54.49  & 58.41     \\
GramNet~\cite{liu2020global} & 51.16  & \cellcolor{lightgray!20}{99.92} & 53.25  & 50.09 & 50.23 & 51.59 & 50.26 & 52.91  & 53.63  & 57.00       \\
RECCE~\cite{Cao_2022_CVPR} & 66.64  & \cellcolor{lightgray!20}{\textbf{100.0}}  & 70.91 & 73.10 & 71.62 & 77.52 & 64.14 & 62.19 & 65.29 & 72.38         \\
LNP~\cite{liu2022detecting} & 80.06 & \cellcolor{lightgray!20}{99.43}  & 67.35 & 50.10 & 54.13 & 50.06 & 50.55 & 50.40 & 50.59 & 61.41  \\
LGrad~\cite{tan2023learning} & 52.94 & \cellcolor{lightgray!20}{\uline{99.99}} & \textbf{99.80} & 64.91 & 57.59 & 66.58 & 60.14 & 76.59  & 74.03  & 72.43         \\
DIRE~\cite{wang2023dire} & 72.48  & 69.81  & \cellcolor{lightgray!20}{\uline{98.92}} & 77.80 & 58.84 & 89.05  & 62.50  & 90.75 & 87.79 & 78.66        \\
Ojha23~\cite{ojha2023towards} & 65.45 & \cellcolor{lightgray!20}{83.40} & 70.06 & 72.25 & 72.76 & 78.55 & 56.21 & 54.96 & 58.01 & 67.96        \\
AEROBLADE~\cite{ricker2024aeroblade} & 48.69 & 51.99 & 69.50 & 42.53 & 46.19 & 90.05 & 77.40 & 82.16 & 81.95 & 65.61       \\
FatFormer~\cite{liu2024forgeryaware} & \textbf{98.91} & \cellcolor{lightgray!20}{98.30} & 97.82 & \uline{95.63}  & 68.88 & 81.04 & \textbf{90.28} & 88.20 & 88.08 & \uline{89.68}        \\
Zou25~\cite{zou2025self} & 76.88 & 74.59 & 93.83 & 93.63 & \uline{78.62} & 95.31 & 83.79 & 91.29 & 91.71 & 86.63 \\
\hline
BLADES-OC (Ours) & 76.75 & 76.78 & 93.63 & \textbf{96.05} & \textbf{80.70} & \textbf{96.09} & \uline{84.82} & \uline{92.79} & \textbf{94.48} & 88.01       \\
BLADES-BC (Ours) & \uline{94.22} & \cellcolor{lightgray!20}{97.24} & 96.95 & 94.33 & 74.83 & \uline{95.73} & 84.40 & \textbf{95.19} & \uline{93.84} & \textbf{91.86}  \\
\bottomrule
\end{tabular}
}
\caption{Cross-generator detection accuracy (\%) of our BLADES-OC and BLADES-BC versus ten competing detectors across nine generative models.
Except for DIRE~\cite{wang2023dire}, AEROBLADE~\cite{ricker2024aeroblade}, Zou25~\cite{zou2025self}, and our OC variant, all detectors are trained on synthetic faces~\cite{esser2021taming} produced by a CelebA-HQ-trained VQGAN (positive class) and CelebA-HQ face photographs~\cite{karras2017progressive} (negative class). DIRE uses the LDM~\cite{rombach2022high} training set; Zou25 and our BLADES-OC rely solely on face photographs; AEROBLADE is training-free. Cells shaded in \colorbox{lightgray!20}{gray} denote results on images by the training generator.  Boldface and underlining denote first‑ and second‑best results, respectively.
}
\label{tab:cross-model}
\end{table*}

\subsection{AI-Generated Face Detection} 
After self-supervised pretraining, we discard the text encoder $\bm f_{\bm \varphi}$ and freeze the vision encoder $\bm f_{\bm \phi}$ as a fixed feature extractor. We explore two detection paradigms.

\noindent\textbf{One-Class Anomaly Detection.} We fit a GMM to embeddings of photographic face images. At test time, any sample whose log-likelihood under this GMM falls below a set threshold (\eg, the $5$-th percentile of the training distribution) is flagged as AI-generated~\cite{zou2025self}.

\noindent\textbf{Binary Classification.} We train a lightweight two-layer perceptron on a small set of photographic and AI-generated image embeddings, using a standard cross-entropy loss. During inference, this classifier outputs the probability of an image being AI-generated~\cite{ojha2023towards, tan2023learning}.

\section{Experiments}
We evaluate the proposed BLADES on a variety of benchmarks to demonstrate its effectiveness and generality for AI‑generated face detection.

\subsection{Experimental Setups}
\noindent\textbf{Datasets.}\label{sec: datasets}
For self-supervised pretraining, we assemble $352,961$ face photographs with rich EXIF metadata from the FDF dataset~\cite{hukkelaas2019deepprivacy} under  CC BY 2.0, filtering nine informative tags\footnote{These include \texttt{aperture}, \texttt{exposure mode}, \texttt{exposure program}, \texttt{exposure time}, \texttt{focal length}, \texttt{ISO speed}, \texttt{makes}, \texttt{metering mode}, and \texttt{white balance mode}.} with $130,000$ faces remaining. For evaluation, we generate synthetic faces using nine state-of-the-art models---StyleGAN2~\cite{karras2020analyzing}, VQGAN~\cite{esser2021taming}, LDM~\cite{rombach2022high}, DDIM~\cite{song2021denoising}, Stable Diffusion 2.1 (SDv2.1)~\cite{rombach2022high}, FreeDoM~\cite{yu2023freedom}, HPS~\cite{wu2023better}, Midjourney~\cite{Midjourney}, and SDXL~\cite{podell2024sdxl}---sourced from public repositories~\cite{esser2021taming, chen2024diffusionface, cheng2024diffusion} or produced on demand. Test face photographs are drawn from the CelebA-HQ~\cite{karras2017progressive} and FFHQ~\cite{karras2019style} datasets.

\noindent\textbf{Implementation Details.}
We adopt ResNet-50~\cite{he2016deep} as the vision encoder and GPT-2~\cite{radford2019GPT-2} as the text encoder within a CLIP-style joint embedding~\cite{radford2021CLIP}. During self-supervised pretraining, the inner loop uses AdamW~\cite{loshchilov2017decoupled} with decoupled weight decay $10^{-3}$, initial learning rate of $10^{-5}$ under cosine annealing~\cite{loshchilov2016sgdr}, minibatch size $48$, and $20$ epochs. The focal-loss parameter $\gamma$ in Eq.~\eqref{eq:focal_cls_exif} is set to $2$~\cite{lin2017focal}. The outer loop initializes all task weights to $1$, which are optimized using Adam~\cite{Kingma2014adam} with a learning rate of $3\times10^{-4}$. Input images are resized to $224\times224\times3$. 
In the one-class classification (OC) setting, we fit a $10$-component GMM with a likelihood threshold that gives a $5\%$ false alarm rate.
In the binary classification (BC) setting, we train a two-layer perceptron (\ie, $768\rightarrow 1,536\rightarrow 2$ with ReLU in between) with Adam~\cite{Kingma2014adam} (learning rate $3\times10^{-4}$ and minibatch size $64$), using low-likelihood face photographs as pseudo-outliers to augment AI-generated faces~\cite{du2022vos}.
All experiments run on a single NVIDIA RTX 3090 GPU.

\noindent\textbf{Evaluation Metrics.} We report the accuracy (Acc), area under the ROC curve (AUC), and average precision (AP) results. Sensitivity and specificity are further analyzed via the true positive rate (TPR), true negative rate (TNR), false positive rate (FPR), false negative rate (FNR), and F‑score.

\begin{figure*}[]
  \centering 
  \subfloat[CLIP~\cite{radford2021CLIP}]{\includegraphics[width=0.49\linewidth]{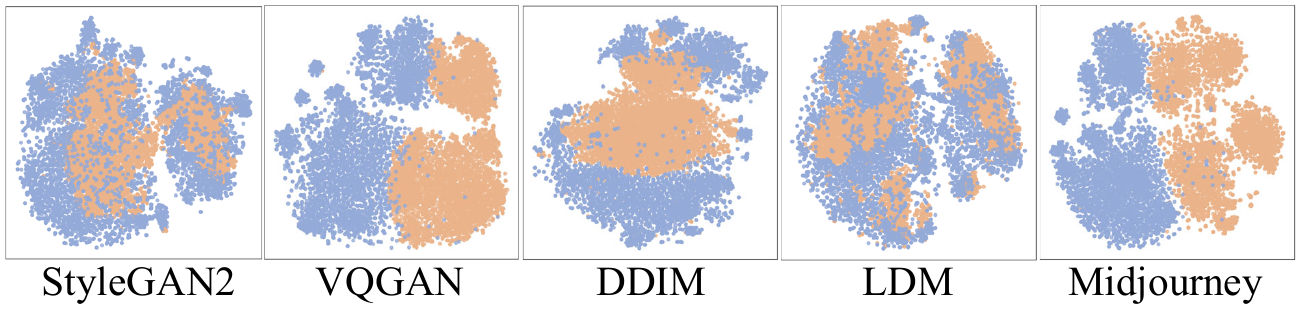}} \hfill
  \subfloat[FaRL~\cite{zheng2022general}]{\includegraphics[width=0.49\linewidth]{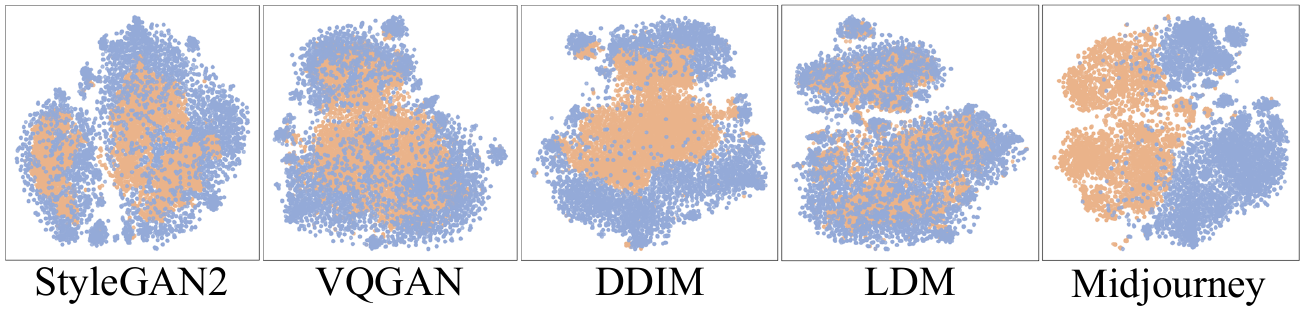}} \hfill
  \subfloat[EAL~\cite{zheng2023exif}]{\includegraphics[width=0.49\linewidth]{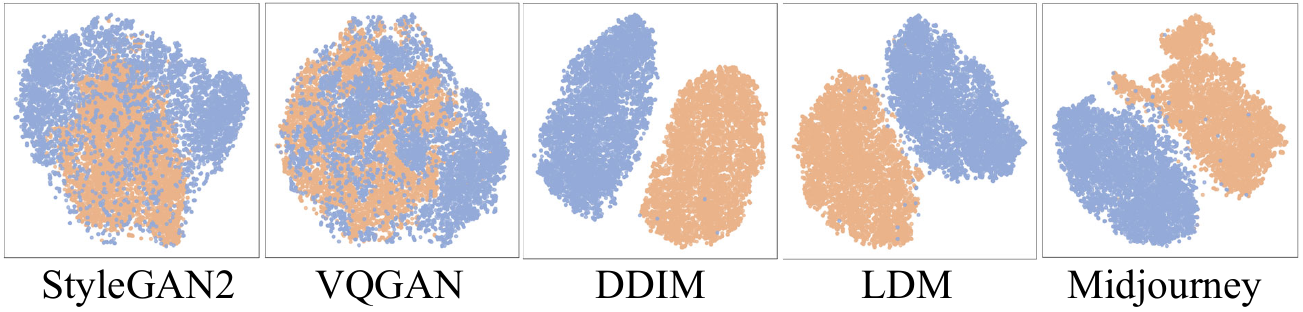}} \hfill
  \subfloat[BLADES-OC]{\includegraphics[width=0.49\linewidth]{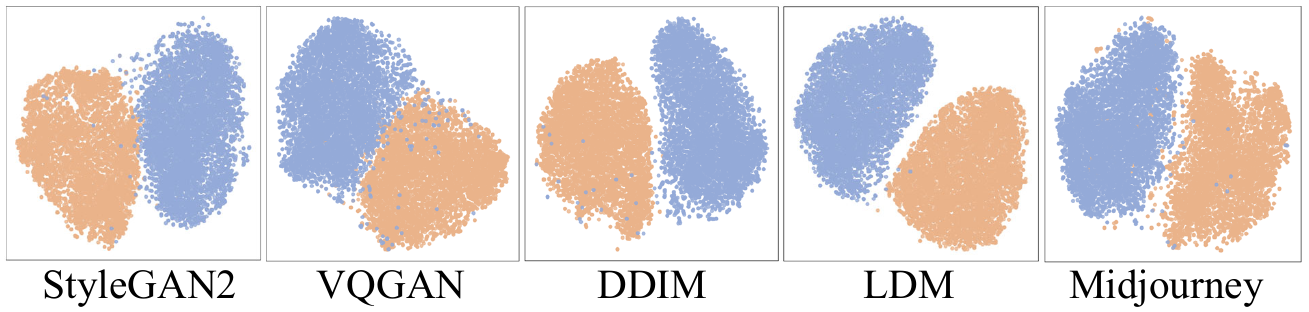}} \hfill
  \caption{t-SNE visualizations~\cite{van2008visualizing} of learned features of photographic (yellow) and AI-generated (blue) face images across five generators.}
  \label{fig: tsne_feature}
\end{figure*}

\begin{table}[]
\centering
\resizebox{\linewidth}{!}{
\begin{tabular}{lccccc}
\toprule
\multirow{2}{*}{Method} & \multicolumn{2}{c}{Photographic (\%)} & \multicolumn{2}{c}{AI-generated (\%)} & \multirow{2}{*}{F-score$\uparrow$}\\
\cmidrule(lr){2-3} \cmidrule(lr){4-5}
& TNR$\uparrow$       & FPR$\downarrow$            & TPR$\uparrow$       & FNR$\downarrow$ &              \\
\hline
LGrad~\cite{tan2023learning} & \textbf{99.98} & \textbf{0.02} & 44.97 & 55.03 & 0.60  \\
DIRE~\cite{wang2023dire} & 99.96 & 0.04 & 54.37 & 45.63 & 0.65  \\
Ojha23~\cite{ojha2023towards} & 75.49 & 24.51 & 60.39 & 39.61 & 0.64 \\
FatFormer~\cite{liu2024forgeryaware} & 97.61 & 2.38 & 81.74 & 18.26 & 0.87  \\
\hline
BLADES-BC & 94.64 & 5.36 & \textbf{88.97} & \textbf{11.03} & \textbf{0.91}  \\
\bottomrule
\end{tabular}
}
\caption{ Sensitivity and specificity analysis. For photographic faces, we report true negative rate (TNR) and false positive rate (FPR); for AI‑generated faces, true positive rate (TPR) and false negative rate (FNR). The F‑score summarizes overall detection performance. Results are averaged across all test generators.
}
\label{tab:confusion_matrix}
\end{table}

\subsection{Main Results}
\noindent\textbf{Cross-Generator Evaluation.} We train detectors using $25,000$ CelebA-HQ face photographs and synthetic face images from a single generator, and then test on the remaining eight. As shown in Table~\ref{tab:cross-model}, our BLADES-BC achieves an average accuracy of $91.86\%$,
outperforming all competing detectors by a substantial margin. Remarkably, our BLADES-OC, trained exclusively on face photographs, surpasses most supervised detectors despite never being exposed to AI-generated faces. This highlights the effectiveness of our bi-level optimization strategy in aligning pretext tasks with the downstream detection objective.
In contrast, GAN-based detectors struggle to generalize to diffusion models, and vice versa. More surprisingly, diffusion-based methods such as DIRE~\cite{wang2023dire} and AEROBLADE~\cite{ricker2024aeroblade} exhibit poor cross-generator generalization even within the same generative family. This suggests that detectors relying on model-specific cues (\eg, reconstruction errors~\cite{wang2023dire, ricker2024aeroblade}) may be inherently fragile as generative models evolve.

\noindent\textbf{Sensitivity and Specificity Analysis.}
Table~\ref{tab:confusion_matrix} breaks down TPR, TNR, FPR, FNR, and F-score across generators. Prior methods exhibit high specificity (with $\text{TNR} > 97\%$) but low sensitivity (with $\text{TPR} < 82\%$), indicating overfitting to seen generators.
In contrast, our BLADES-BC balances both, achieving $94.64\%$ TNR and $88.97\%$ TPR (with an F-score of $0.91$), reflecting robust cross-generator detection.

\noindent\textbf{Cross-Dataset Evaluation.} To assess domain transfer, we train on CelebA-HQ~\cite{karras2017progressive} and evaluate  using 
FFHQ~\cite{karras2019style} face photographs and FFHQ-trained generators (\ie, StyleGAN2~\cite{karras2019style}, VQGAN~\cite{esser2021taming}, and LDM~\cite{rombach2022high}).
As reported in Table~\ref{tab:cross-dataset}, the proposed BLADES-BC consistently achieves strong performance, surpassing GAN-based and diffusion-based detectors. These results confirm that BLADES' self-supervised features generalize across photographic domains and synthesis pipelines.

\begin{table}[]
\setlength\tabcolsep{2.5pt}
\centering
\resizebox{\linewidth}{!}{
\begin{tabular}{lcccccccc}
\toprule
\multirow{2}{*}{Method} & \multicolumn{2}{c}{StyleGAN2} & \multicolumn{2}{c}{VQGAN} & \multicolumn{2}{c}{LDM} & \multicolumn{2}{c}{Average} \\
\cmidrule(lr){2-3} \cmidrule(lr){4-5} \cmidrule(lr){6-7} \cmidrule(lr){8-9} 
& AUC            & AP           & AUC          & AP         & AUC         & AP        & AUC           & AP          \\
\hline
LGrad~\cite{tan2023learning}     & 79.16 & 81.43 &  \textbf{99.80} & \textbf{99.97} & \textbf{97.09} & \textbf{97.13} & 92.02 & 92.84 \\
DIRE~\cite{wang2023dire}      & 81.57 & \textbf{94.48} & 78.95 & 94.27 & 94.03 & 93.99 & 84.85 & 94.25 \\
Ojha23~\cite{ojha2023towards} & \textbf{90.38} & 88.62 & 96.45 & 96.83 & 90.90 & 91.16 & 92.58 & 92.20      \\
FatFormer~\cite{liu2024forgeryaware} & 87.93 & 85.70 & 97.69 & 97.34 & 94.85 & 93.93 & 93.49 & 92.32      \\
\hline
BLADES-BC   &     88.16      &   87.96    &  98.52   &  98.54 & 96.19 & 96.14 & \textbf{94.29} & \textbf{94.21}       \\
\bottomrule
\end{tabular}
}
\caption{Cross-dataset detection results (\%). All detectors are trained on CelebA-HQ~\cite{karras2017progressive} and CelebA-HQ-trained generators, and evaluated using FFHQ~\cite{karras2019style} face photographs and FFHQ-trained generators.
% , except for DIRE~\cite{wang2023dire}, which is trained on LDM-generated faces~\cite{rombach2022high}.
}
\label{tab:cross-dataset}
\end{table}

\subsection{Further Analysis}
\noindent\textbf{Feature Separability Analysis.} To assess the effectiveness of our learned representations, we compare their separability against those produced by existing one-class detectors---Hu21~\cite{hu2021exposing}, LNP~\cite{liu2022detecting}, and Zou25~\cite{zou2025self}, as well as general-purpose visual features from  CLIP~\cite{radford2021CLIP}, FaRL~\cite{zheng2022general}, and EXIF-as-language (EAL)~\cite{zheng2023exif}.

We conduct one-class classification using a GMM as the default anomaly detector, except for Hu21 and LNP, which follow their original classification protocols. As shown in Table~\ref{tab:pretrained_feature}, our method achieves the highest average AUC across nine generators, outperforming all competitors by a clear margin. The handcrafted features of Hu21 perform poorly, likely due to the unrealistic assumption of visible corneal specular highlights. Learning-based features show better separability~\cite{liu2022detecting, radford2021CLIP, zheng2022general, zheng2023exif, zou2025self}, with EXIF-induced EAL~\cite{zheng2023exif} and Zou25~\cite{zou2025self} demonstrating strong performance. Our superior results validate the design of our EXIF-based pretext tasks and further highlight the advantage of aligning pretraining objectives with downstream detection via bi-level optimization.

\begin{table*}[]
\small
\centering
\resizebox{0.9\linewidth}{!}{
\begin{tabular}{lcccccccccc}
\toprule
Method & StyleGAN2 & VQGAN & LDM & DDIM & SDv2.1 & FreeDoM & HPS & Midjourney & SDXL & Average \\
\midrule
CLIP~\cite{radford2021CLIP}     & 33.99          & 60.20          & 55.49          & 82.85          & 90.15          & 85.39          & 91.22          & 92.21          & 93.66          & 76.13 \\
FaRL~\cite{zheng2022general}     & 34.35          & 55.91          & 47.26          & 85.10          & \textbf{95.24} & 79.91          & \textbf{93.97} & 89.72          & 94.61          & 75.12 \\
EAL~\cite{zheng2023exif}      & 69.71          & 72.41          & 85.81          & 84.98          & 74.21          & 97.92          & 87.33          & 91.65          & 93.04          & 84.12 \\
Hu21~\cite{hu2021exposing}  & 50.00          & 49.99          & 49.99          & 49.99          & 49.99          & 49.99          & 49.99          & 50.00          & 49.99          & 49.99 \\
LNP~\cite{liu2022detecting}      & 37.55          & 63.28          & 71.12          & 69.54          & 65.64          & 66.40          & 66.64          & 66.70          & 67.25          & 63.79 \\
Zou25~\cite{zou2025self}    & 85.69          & 84.31          & \textbf{98.66} & 97.96          & 88.29          & 99.79          & 91.92          & 97.07          & 97.23          & 93.43 \\
\hline
BLADES-OC & \textbf{87.89} & \textbf{88.31} & 98.24          & \textbf{99.32} & 91.72          & \textbf{99.92} & 93.93          & \textbf{97.80} & \textbf{98.36} & \textbf{95.05} \\
\bottomrule
\end{tabular}
}
\caption{Feature separability comparison in terms of AUC (\%).}
\label{tab:pretrained_feature}
\end{table*}

\begin{table*}[]
\small
\centering
\resizebox{0.9\linewidth}{!}{
\begin{tabular}{lcccccccccc}
\toprule
Method & StyleGAN2 & VQGAN & LDM & DDIM & SDv2.1 & FreeDoM & HPS & Midjourney & SDXL & Average \\
\midrule
FatFormer~\cite{liu2024forgeryaware} & \textbf{85.48} & 93.20 & \textbf{88.41} & 87.78 & 48.77 & 91.95 & 65.65 & 73.48 & 70.23 & 78.33 \\

BLADES-BC & 73.10 & \textbf{96.87} & 71.54 & \textbf{96.10} & \textbf{63.35} & \textbf{95.94} & \textbf{81.70} & \textbf{96.97} & \textbf{96.79} & \textbf{85.82} \\

\bottomrule
\end{tabular}
}
\caption{AUC results (\%) of the proposed BLADES-BC against FatFormer under mild JPEG compression.}
\label{tab:jpeg_robust}
\end{table*}

\begin{figure}[]
  \centering
  \includegraphics[width=0.8\linewidth]{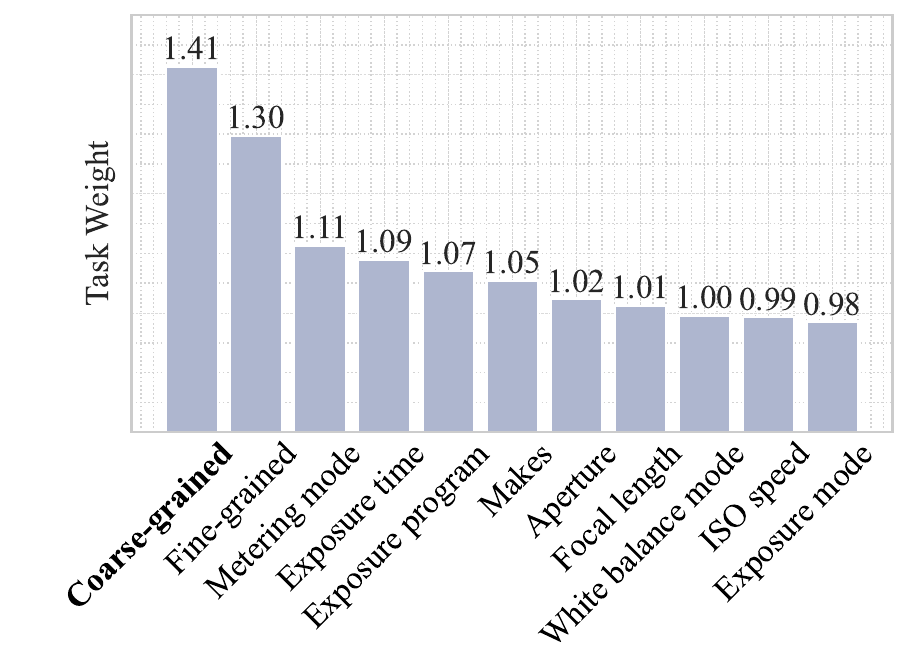}
  \caption{Learned loss weights across pretext tasks, where higher values indicate greater relevance to the primary task (in bold).}
  \label{fig: task_relations}
\end{figure}

Fig.~\ref{fig: tsne_feature} provides t-SNE visualizations~\cite{van2008visualizing} of feature embeddings of photographic (yellow) and AI-generated (blue) faces. Our features yield the most distinct separation, indicating high discriminative capability. In contrast, CLIP and FaRL features exhibit greater overlap, suggesting limited suitability of these semantic representations for AI-generated face detection.

\noindent\textbf{Task Relationship Analysis.}
We investigate the relationships among pretext tasks by analyzing their learned weights from bi-level optimization. As shown in Fig.~\ref{fig: task_relations}, these weights, sorted in descending order, offer several insights into task relevance. First, coarse-grained face manipulation detection---used as the surrogate primary task---receives the highest weight. This confirms its pivotal role in aligning the self-supervised feature space with the downstream goal of AI-generated face detection. Its fine-grained counterpart, which focuses on detecting localized face manipulations, is also assigned substantial importance, indicating that both global and local perturbations provide valuable supervisory signals. Second, among the EXIF-related tasks, those associated with exposure settings, including \texttt{metering mode}, \texttt{exposure time}, \texttt{exposure program}, and \texttt{aperture}, are weighted more heavily than others. This suggests that exposure-related metadata may capture subtle statistical cues that differentiate photographic and AI-generated faces.

To assess the discriminative power of the most influential tasks, we retrain BLADES using only the top six pretext tasks ranked by their weights. This streamlined configuration achieves competitive one-class detection performance ($87.62\%$ mAcc and $94.80\%$ mAP), only marginally lower than the full-task model. These results demonstrate that bi-level optimization not only identifies but also prioritizes the most informative self-supervised objectives for AI-generated face detection.

\begin{table}
\small
\centering
\resizebox{\linewidth}{!}{
\begin{tabular}{lccc}
\toprule
Method    & \#Params (M) & FLOPs (M) & Inference time (s) \\
\hline
FatFormer~\cite{liu2024forgeryaware} & 202  & 51,895             & 0.1                \\
BLADES-BC & 26   & 4,134 & 0.005 \\   
\bottomrule
\end{tabular}
}
\caption{Model complexity comparison during inference. We report the model parameter count, total number of floating-point operations (FLOPs), and inference time per $224\times 224\times 3$ image on an NVIDIA RTX 3090 GPU.}
\label{tab: model_complexity}
\end{table}

\noindent\textbf{Robustness Analysis.} We assess the robustness of the proposed BLADES-BC under mild JPEG compression (quality factor $95$), a common post-processing operation that may obscure low-level statistical cues. As shown in Table~\ref{tab:jpeg_robust}, BLADES-BC consistently outperforms FatFormer across most generators, indicating stronger resilience of our self-supervised features to compression-induced artifacts.
Nonetheless, we see noticeable performance degradations (\eg, accuracy drops from $91.86\%$ to $78.87\%$).

\noindent\textbf{Model Complexity Analysis.} We further compare the computational efficiency of BLADES-BC against FatFormer~\cite{liu2024forgeryaware} in terms of the model parameter count, total number of floating-point operations (FLOPs),  and inference time in Table~\ref{tab: model_complexity}. BLADES-BC achieves a favorable complexity profile, requiring $8\times$ fewer parameters and $12\times$ fewer FLOPs, and running $20\times$ faster during inference. This demonstrates that BLADES-BC is well-suited for deployment in resource-constrained or latency-sensitive applications, without sacrificing detection performance.

%-------------------------------------------------------------------------
\begin{table*}[t]
\centering
\captionsetup{font=small}
    \begin{subtable}[t]{0.43\textwidth}
        \centering
        \renewcommand\arraystretch{1.0}
        \caption{Model structure ablation.}
        \resizebox{0.78\linewidth}{!}{
        \begin{tabular}{cc|cc} 
        \toprule
        Training $\bm f_{\bm \varphi}$ & Joint embedding & mAcc & mAP \\
        \hline
        \xmark & \xmark & 85.93 & 93.44 \\
        \xmark & \cmark & 81.65 & 91.89 \\
        \cellcolor{lightgray!20}{\cmark}   & \cellcolor{lightgray!20}{\cmark} & \cellcolor{lightgray!20}{\textbf{88.01}} & \cellcolor{lightgray!20}{\textbf{95.24}} \\
        \bottomrule
        \end{tabular}
        }
        \label{tab:ablation_structure}
    \end{subtable}
    \hfill
    \begin{subtable}[t]{0.55\textwidth}
        \centering
        \renewcommand\arraystretch{1.0}
        \caption{Training strategy ablation.}
        \resizebox{0.9\linewidth}{!}{
        \begin{tabular}{ccc|cc}
         \toprule
        Focal weighting & Pairwise ranking & Bi-level optimization & mAcc & mAP \\
        \hline
        \cmark & \cmark &  \xmark & 86.20  & 93.78  \\
        \cmark & \xmark & \cmark & 81.52  & 93.14  \\
        \xmark & \cmark & \cmark &  81.87 & 93.14  \\
        \cellcolor{lightgray!20}{\cmark}   & \cellcolor{lightgray!20}{\cmark}   & \cellcolor{lightgray!20}{\cmark} & \cellcolor{lightgray!20}{\textbf{88.01}} & \cellcolor{lightgray!20}{\textbf{95.24}} \\
        \bottomrule
        \end{tabular}
        }
        \label{tab:ablation_optim} 
     \end{subtable}
      \hfill
     \begin{subtable}[t]{0.55\textwidth}
        \centering
        \renewcommand\arraystretch{1.0}
        \vspace{2mm}
        \caption{Primary task ablation.}
        \resizebox{0.95\linewidth}{!}{
        \begin{tabular}{l|cc}
         \toprule
        Primary task configuration & mAcc & mAP \\
        \hline
        Binary classification of photographic vs. AI-generated faces & 86.75  & 92.66  \\
        Fine-grained face manipulation detection & 87.11  & 95.02  \\
        Equal weighting over all pretext tasks & 87.84  & 94.57  \\
        \cellcolor{lightgray!20}{Coarse-grained face manipulation detection} & \cellcolor{lightgray!20}{\textbf{88.01}} & \cellcolor{lightgray!20}{\textbf{95.24}} \\
        \bottomrule
        \end{tabular}
        }
        \label{tab:ablation_bi_level} 
     \end{subtable}
    \hfill
    \begin{subtable}[t]{0.43\textwidth}
        \centering
        \renewcommand\arraystretch{1.0}
        \vspace{2mm}
        \caption{Pretext task ablation.}
        \resizebox{0.8\linewidth}{!}{
        \begin{tabular}{ccc|cc}
        \toprule
         EXIF   & Coarse-grained & Fine-grained  & mAcc & mAP \\
        \hline
        \cmark & \xmark & \xmark & 81.80 & 92.13  \\
        \xmark & \cmark & \cmark & 86.49 & 93.40  \\
        \cmark & \cmark & \xmark & 87.03 & 94.72  \\
        \cellcolor{lightgray!20}{\cmark}   & \cellcolor{lightgray!20}{\cmark}   & \cellcolor{lightgray!20}{\cmark} & \cellcolor{lightgray!20}{\textbf{88.01}} & \cellcolor{lightgray!20}{\textbf{95.24}} \\
        \bottomrule
        \end{tabular}}
        \label{tab:ablation_tasks}
    \end{subtable}
\caption{Ablation studies of key design choices in BLADES for self-supervised pretraining. The default setting is shaded \colorbox{lightgray!20}{gray}.
}
\label{tab: ablations_all}
\end{table*}
%-------------------------------------------------------------------------

\subsection{Ablation Studies}

\noindent\textbf{Model Structure.} We examine the importance of joint embedding and text encoder training. One variant omits joint embedding entirely, while another freezes the text encoder during pretraining. Table~\ref{tab:ablation_structure} demonstrates that co-optimizing the vision and text encoders significantly enhances performance, validating the necessity of a fully trainable implementation of joint embedding. 

\noindent\textbf{Training Strategy.} We assess the contributions of three core strategies: focal loss weighting to mitigate class imbalance (in Eq.~\eqref{eq:focal_cls_exif}), pairwise ranking of ordinal tags (in Eqs.~\eqref{eq: weightes_sum} to~\eqref{eq: fidelity}), and bi-level optimization for dynamic task weighting (in Eq.~\eqref{eq:auto-weighting}).
Table~\ref{tab:ablation_optim} shows that disabling any component\footnote{Disabling focal weighting is done by setting $\gamma = 0$ in Eq.~\eqref{eq:focal_cls_exif}; disabling pairwise ranking corresponds to direct alignment between images and ordinal EXIF tags as done in~\cite{zheng2023exif}; disabling bi-level optimization refers to training BLADES using only Eq.~\eqref{eq: all_tasks} with equal weights.} leads to performance drops, with the full configuration achieving the highest accuracy and precision.

\noindent\textbf{Primary Task Configuration.} We investigate different choices for the outer-loop primary task in Eq.~\eqref{eq:primary}, including 1) binary classification of photographic vs. VQGAN-generated faces~\cite{esser2021taming},  
2) fine-grained face manipulation detection, 3) equal weighting over all pretext tasks, and 4) our default setting---coarse-grained face manipulation detection. Table~\ref{tab:ablation_bi_level} indicates that our default setting yields the best results, while binary classification of photographic vs. AI-generated faces underperforms, likely due to overfitting on synthesis artifacts.

\noindent\textbf{Pretext Task Composition.} We further evaluate the impact of different pretext task combinations. Variants exclude one or more tasks from the full setting, specifically, coarse- and fine-grained face manipulation detection and EXIF-based tasks.  
Table~\ref{tab:ablation_tasks} shows that excluding any task degrades performance. The superiority of the full combination shows that EXIF metadata and face manipulations provide complementary cues for robust AI-generated face detection.

\section{Conclusion and Discussion}
In this work, we have introduced BLADES for detecting AI-generated face images. Our method aligns self-supervised feature learning with the downstream detection task by dynamically weighting a suite of EXIF- and manipulation-based pretext objectives via bi-level optimization. Once trained, the learned representations are shown to be highly discriminative in both one-class and binary classification settings, achieving state-of-the-art generalization performance across diverse generative models and datasets.

While our focus has been on AI-generated face detection, the underlying principles of BLADES, especially the task-aware self-supervision and bi-level alignment, are broadly applicable. A promising direction for future research is to generalize these ideas to broader natural scenes, where forged content may be subtler and less structurally constrained. This transition demands new surrogate tasks beyond face manipulations, possibly involving semantic consistency, scene dynamics, or temporal coherence.

Another important avenue is to enhance robustness to benign post-processing, such as mild compression, rescaling, or filtering. This could be addressed by reducing over-reliance on high-frequency cues, encouraging the modeling of perturbation-invariant features, or incorporating contrastive or augmentation-aware pretext losses during self-supervised pretraining.

Finally, we envision enriching interpretability and decision-making through the integration of multimodal large language models. These models can be used not only to describe visual anomalies in natural language but also to reason about \textit{imperceptible} artifacts, provide explanations for decision-making, and even invoke external tools or knowledge bases to justify their predictions. Such capabilities would elevate BLADES from a powerful detector to a transparent and explainable forensic assistant, particularly valuable in high-stakes scenarios involving legal, journalistic, or national security contexts.

\section*{Acknowledgements}
This work was supported in part by the Hong Kong RGC General Research Fund (11220224), the CityU Strategic Research Grants (7005848 and 7005983), and an Industry Gift Fund (9229179).

{
    \small
    \bibliographystyle{ieeenat_fullname}
    \bibliography{DeepFake-25-ICCV}
}

\clearpage
\maketitle
\maketitlesupplementary
\appendix

\section{Computation of Bi-level Optimization}
\label{supp_sec: computation_blo}

In solving the bi-level optimization Problem~\eqref{eq:auto-weighting}, parameters $\bm \theta$ and $\bm \lambda$ are updated alternately:
\begin{subequations}\label{supp_eq: param_update}
    \begin{align}
        \bm \theta' &= \bm \theta - \alpha \nabla_{\bm \theta} \sum_{\bm x\in \mathcal{B}_{\mathrm{tr}}} \sum_{i=1}^K \lambda_i \ell_i(\bm x; \bm \theta) \label{supp_subeq:theta'} \\
        \bm \lambda &\leftarrow \bm \lambda - \beta \nabla_{\bm \lambda} \sum_{\bm x\in \mathcal{B}_{\mathrm{val}}} \ell_1\left(\bm x; \bm \theta'(\bm \lambda)\right) \label{supp_subeq:lambda_update} \\
        \bm \theta &\leftarrow \bm \theta - \alpha \nabla_{\bm \theta} \sum_{\bm x\in \mathcal{B}_{\mathrm{tr}}} \sum_{i=1}^K \lambda_i \ell_i(\bm x; \bm \theta),\label{supp_subeq:theta_update}
    \end{align}
\end{subequations}
where $\alpha$ and $\beta$ are the inner- and outer-loop learning rates, respectively.
To compute $\nabla_{\bm \lambda} \ell_1(\bm x; \bm \theta'(\bm \lambda))$, we apply the chain rule, yielding
\begin{align}\label{supp_eq:lambda_grad}
    \nabla_{\bm \lambda} \ell_1\left(\bm x; \bm \theta'(\bm \lambda)\right) 
    = \nabla_{\bm \lambda} \bm \theta'(\bm \lambda) \nabla_{\bm \theta'} \ell_1\left(\bm x; \bm \theta'(\bm \lambda)\right) .
\end{align}
Making use of Eq.~\eqref{supp_subeq:theta'} and defining $\ell_\mathrm{pre}(\bm \theta, \bm \lambda) = \sum_{\bm x\in \mathcal{B}_{\mathrm{tr}}} \sum_{i=1}^K \lambda_i \ell_i(\bm x; \bm \theta)$, we have the Jacobian:
\begin{align}\label{supp_eq:theta'_lambda_grad}
    \nabla_{\bm \lambda} \bm \theta'(\bm\lambda) 
    &= \nabla_{\bm \lambda} \left( \bm \theta - \alpha \nabla_{\bm \theta} \ell_\mathrm{pre}(\bm \theta, \bm \lambda) \right) \\
    &= -\alpha \nabla^2_{\bm \theta, \bm \lambda} \ell_\mathrm{pre}(\bm \theta, \bm \lambda)^\intercal. \nonumber
\end{align}
Substituting it into Eq.~\eqref{supp_eq:lambda_grad}, we obtain the final expression:
\begin{align}\label{supp_eq:lambda_grad_combine}
    \nabla_{\bm \lambda} \ell_1\left(\bm x; \bm \theta'(\bm \lambda)\right) =- \alpha \nabla^2_{\bm \theta, \bm \lambda} \ell_\mathrm{pre}(\bm \theta,\bm \lambda) ^\intercal
    \nabla_{\bm \theta'} \ell_1\left(\bm x; \bm \theta'(\bm \lambda)\right).
\end{align}

\noindent\textbf{Finite Difference Approximation.} Direct computation of the mixed second-order derivative $\nabla^2_{\bm \theta, \bm \lambda}\ell_\mathrm{pre}(\bm \theta, \bm \lambda)$ is computationally intensive. To circumvent this, we employ a finite difference scheme~\cite{finn2017model_supp, liu2018darts_supp} to approximate the Jacobian-vector product required in the outer-loop gradient.
Let
$\bm v= \nabla_{\bm {\theta'}} \ell_1(\bm x; \bm \theta'(\bm \lambda))$ denotes the gradient of the outer-loop loss with respect to the updated parameters. Define the gradient of the inner-loop objective with respect to $\bm \lambda$ as a function of $\bm \theta$, \ie, $\bm F(\bm \theta)=\nabla_{\bm \lambda} \ell_{\mathrm{pre}}(\bm \theta, \bm \lambda)$. We apply a second-order central difference approximation of the directional derivative of $\bm F$ along $\bm v$:
\begin{align}
    &\nabla^2_{\bm \theta, \bm \lambda}\ell_\mathrm{pre}(\bm \theta,\bm \lambda) ^\intercal \nabla_{\bm {\theta'}}\ell_1(\bm x; \bm \theta'(\bm \lambda))\nonumber\\ 
    & =  \nabla_{\bm \theta} F(\bm \theta)^\intercal \bm v\nonumber\\
    & \approx \frac{\bm F(\bm \theta + \epsilon \bm v)-\bm F(\bm \theta - \epsilon \bm v)}{2\epsilon}\nonumber\\
    &=\frac{\nabla_{\bm \lambda} \ell_{\mathrm{pre}}(\bm \theta^{+}, \bm \lambda)-\nabla_{\bm \lambda} \ell_{\mathrm{pre}}(\bm \theta^{-}, \bm \lambda)}{2\epsilon},
\end{align}
where 
\begin{align}
    \bm \theta^{\pm} = \bm \theta \pm \epsilon \bm v,
\end{align}
and  $\epsilon$ is a small constant.

\begin{table*}[]
\small
\centering
\begin{tabular}{llcc}
\toprule
EXIF tag            & Example value                                     & \#Unique entries & Count \\
\hline
\texttt{Aperture}            & \texttt{F2.8, F4, F5.6, F3.5}                               & 152      & 198,448       \\
\texttt{Exposure Mode}       & \texttt{Auto, Auto-bracketing, Program, Manual}             & 7      & 194,666       \\
\texttt{Exposure Program}    & \texttt{Manual Control, Normal Program, Portrait Mode}      & 6       & 176,787       \\
\texttt{Exposure Time}       & \texttt{1/60 sec, 1/125 sec, 1/250 sec}                     & 1,745     & 198,488       \\
\texttt{Focal Length}        & \texttt{18.0 mm, 50.0 mm, 6.3 mm}                           & 858      & 198,488       \\
\texttt{ISO Speed }          & \texttt{100, 200, 400, 800}                                 & 269      & 198,488       \\
\texttt{Makes}               & \texttt{Canon, Apple, Sony, Nikon}                          & 10      & 198,247       \\
\texttt{Metering Mode}       & \texttt{Center-weighted, Average, Partial, Spot}            & 8       & 196,340       \\
\texttt{White Balance Mode}  & \texttt{Auto, Manual}                              & 2        & 193,279       \\
\hline
\texttt{Custom Rendered}     & \texttt{Custom Process, Normal Process, Unknown}            & 3        & 180,667       \\
\texttt{Date/Time}           & \texttt{2013:03:28 04:20:46}                                & 96,568    & 196,639       \\
\texttt{Date/Time Digitized} & \texttt{2013:03:28 04:20:46}                                & 96,447    & 197,521       \\
\texttt{Date/Time Original}  & \texttt{2013:03:28 04:20:46}                                & 96,827   & 198,227       \\
\texttt{EXIF Version}        & \texttt{2.21, 2.20, 2.30}                                   & 11       & 197,673       \\
\texttt{Flash}               & \texttt{Unfired, Fired, Unknown, Fired Auto}                & 4        & 198,317       \\
\texttt{F-Number}            & \texttt{F2.8, F4, F5.6, F3.5}                               & 111      & 197,501       \\
\texttt{Resolution Unit}     & \texttt{Inch, Cm, No Unit, Unknown}                         & 4        & 193,347       \\
\texttt{Scene Capture Type}  & \texttt{Standard, Portrait, Nightscene, Landscape, Unknown} & 5        & 192,638       \\
\texttt{Shutter Speed}       & \texttt{1/60 sec, 1/125 sec, 1/250 sec}                     & 1,195     & 196,970       \\
\texttt{X Resolution}        & \texttt{72 dots per inch}                                   & 107      & 193,173       \\
\texttt{Y Resolution}        & \texttt{72 dots per inch}                                   & 108      & 193,173       \\
\bottomrule
\end{tabular}
\vspace{-.2cm}
\caption{Overview of EXIF tags appearing in more than $50\%$ of the collected photographic face images. The upper section lists the nine tags retained for pretraining based on relevance, informativeness, and non-redundancy. 
}
\label{supp_tab:exif}
\end{table*}

\section{EXIF Tag Selection}
\label{supp_sec: EXIF_tags}
This section details the procedure for selecting EXIF tags used in our self-supervised pretraining.

We begin by identifying all EXIF tags that appear in more than $50\%$ of the collected photographic face images from the FDF dataset~\cite{zheng2022general}. Table~\ref{supp_tab:exif} lists their frequencies and representative values. To ensure that the selected tags offer meaningful supervisory signals, we apply three empirical filtering criteria:
\begin{itemize}
    \item \textbf{Relevance to Digital Imaging:} The chosen tag must encode semantically or physically interpretable imaging attributes (\eg, exposure settings and camera make).
    \item \textbf{Information Richness:} Tags dominated by unknown entries or a single category are excluded to avoid degenerate supervision.
    \item \textbf{Semantic Redundancy Removal:} Tags whose semantics largely overlap with others (\eg, \texttt{F-number} with \texttt{aperture}) are removed for parsimony.
\end{itemize}
For example, tags such as \texttt{date/time}, \texttt{EXIF version}, \texttt{resolution unit}, and \texttt{scene capture type} are excluded due to irrelevance to image formation. Likewise,  \texttt{flash} is excluded because approximately $\sim\!75\%$ of its values are missing or unknown.  

After refinement, we retain nine EXIF tags for pretraining:  
\texttt{aperture}, \texttt{exposure mode}, \texttt{exposure program}, \texttt{exposure time}, \texttt{focal length}, \texttt{ISO speed}, \texttt{makes}, \texttt{metering mode}, and \texttt{white balance mode}.

\begin{figure*}[]
  \centering
  \subfloat[CelebA-HQ]{\includegraphics[width=\linewidth]{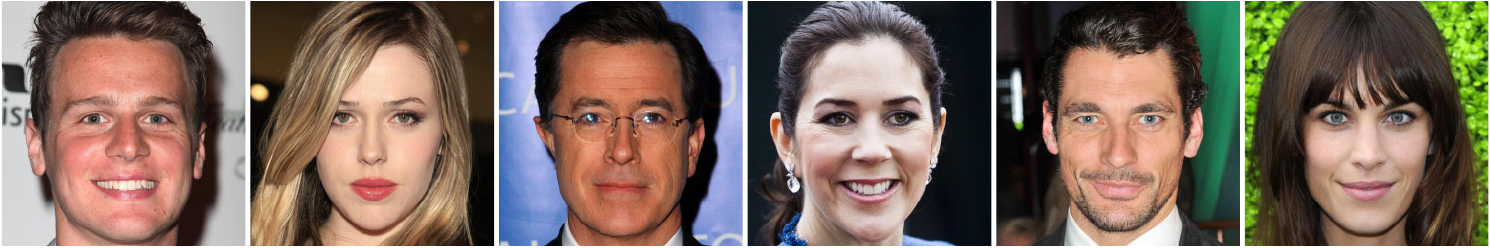}} \\
  \subfloat[StyelGAN2]{\includegraphics[width=\linewidth]{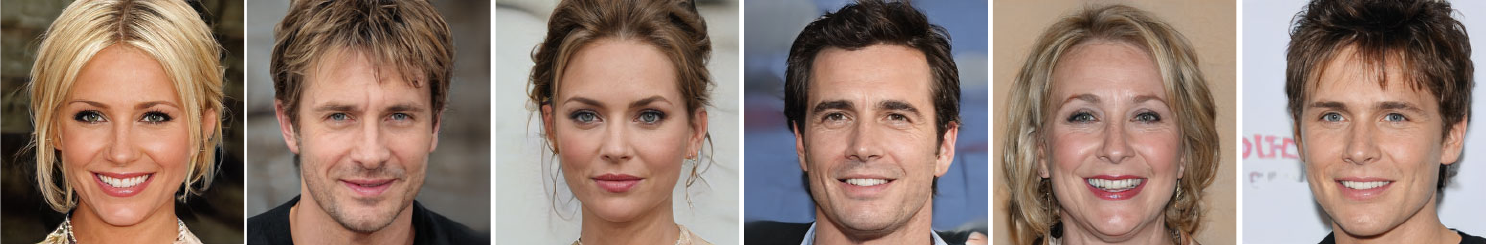}} \\
  \subfloat[VQGAN]{\includegraphics[width=\linewidth]{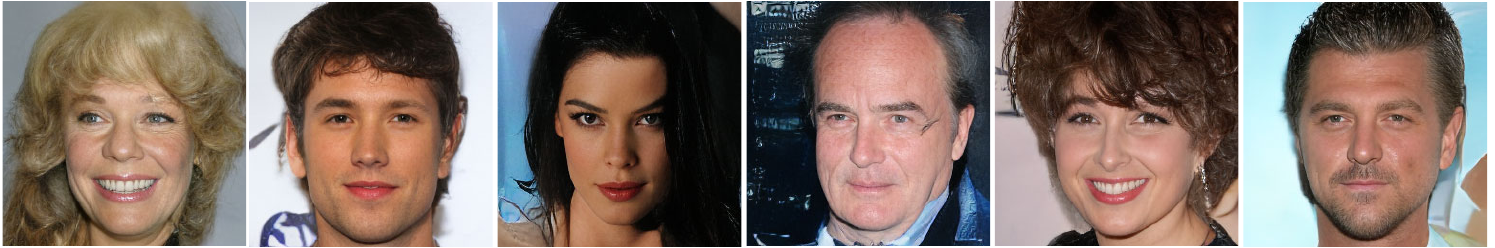}} \\
  \subfloat[LDM]{\includegraphics[width=\linewidth]{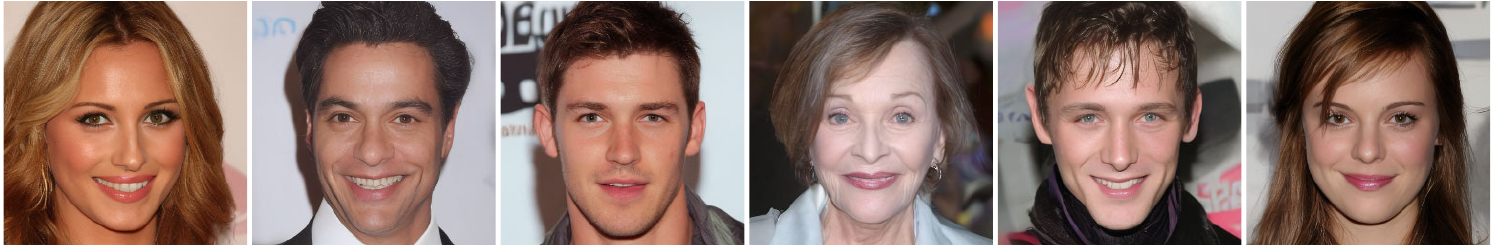}} \\
  \subfloat[DDIM]{\includegraphics[width=\linewidth]{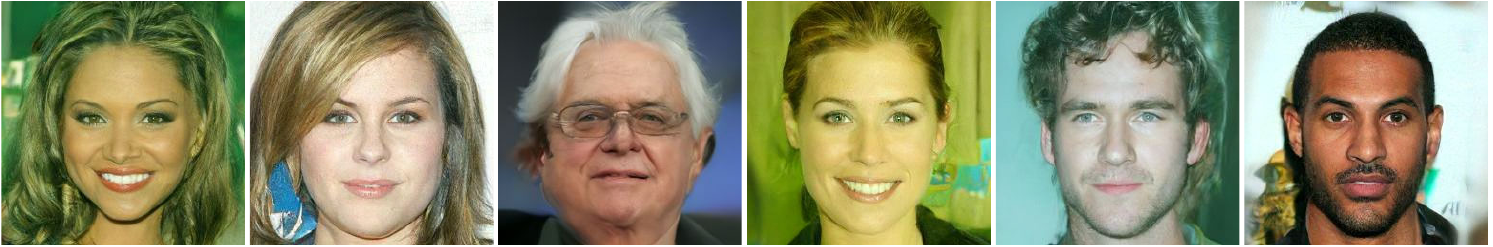}} \\
  \subfloat[SDv2.1]{\includegraphics[width=\linewidth]{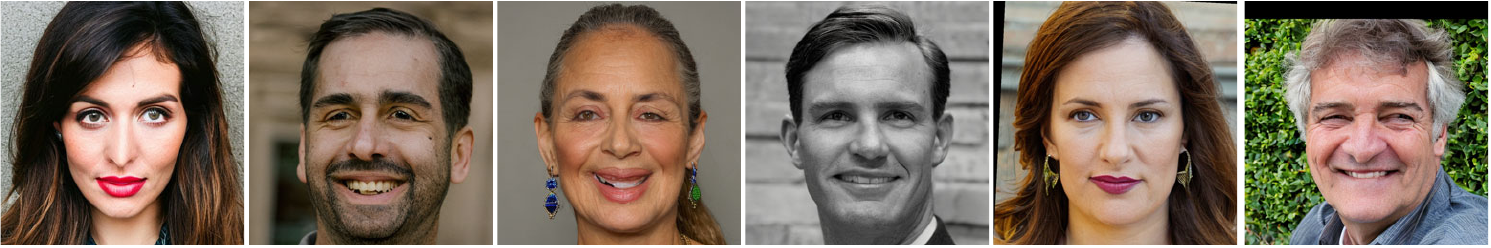}} \\
  \caption{Representative face images used in cross-generator evaluation. (Part 1 of 2).}
  \label{supp_fig:face_cross_model_1}
\end{figure*}

\begin{figure*}[]
  \ContinuedFloat
  \centering
  \subfloat[FreeDoM]{\includegraphics[width=\linewidth]{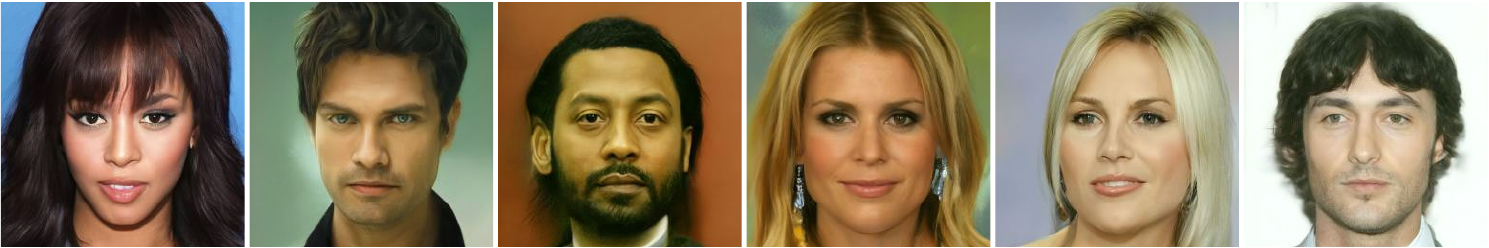}} \\
  \subfloat[HPS]{\includegraphics[width=\linewidth]{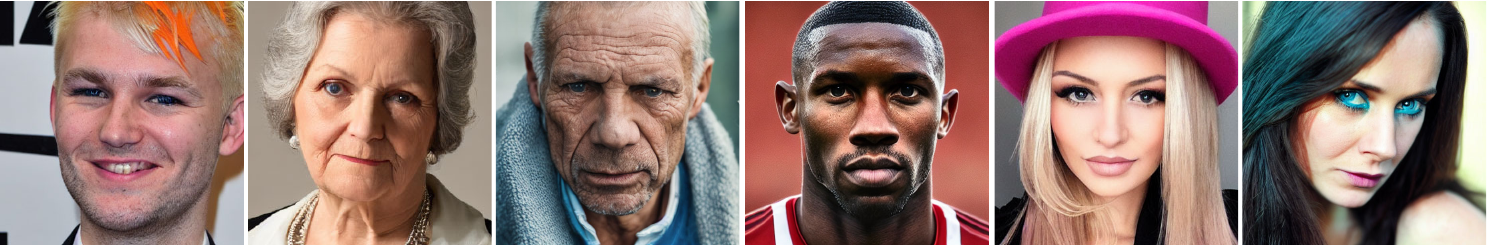}} \\
  \subfloat[Midjourney]{\includegraphics[width=\linewidth]{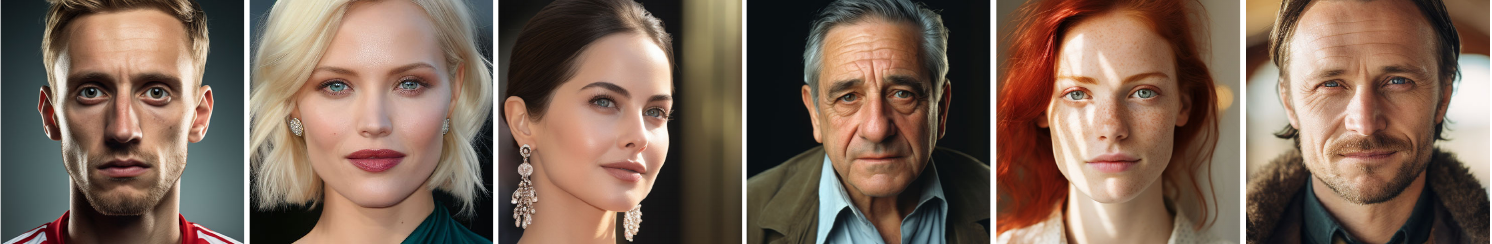}} \\
  \subfloat[SDXL]{\includegraphics[width=\linewidth]{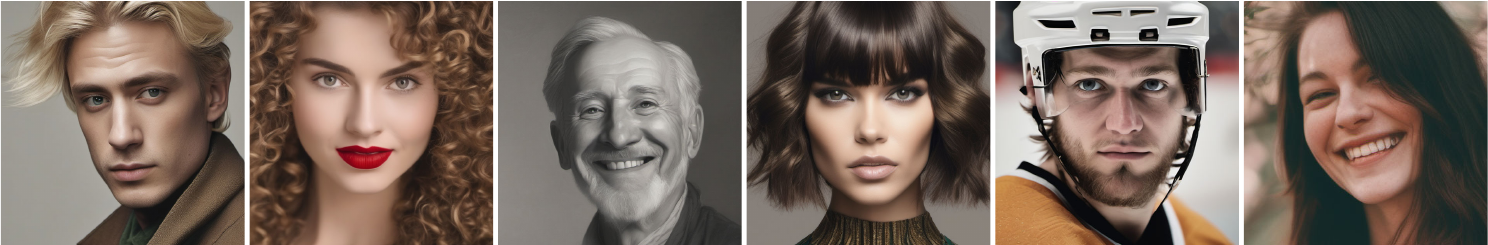}}
  \caption{Representative face images used in cross-generator evaluation. (Part 2 of 2).}
  \label{supp_fig:face_cross_model_2}
\end{figure*}

\begin{figure*}[]
  \centering
  \subfloat[FFHQ]{\includegraphics[width=\linewidth]{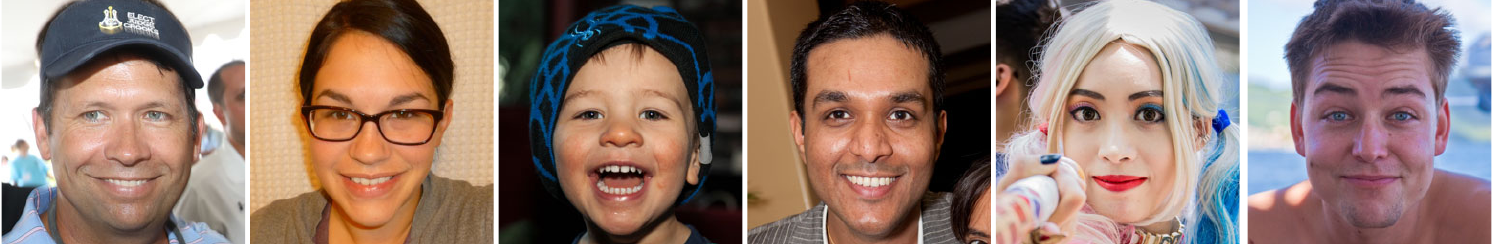}} \\
  \subfloat[StyleGAN2]{\includegraphics[width=\linewidth]{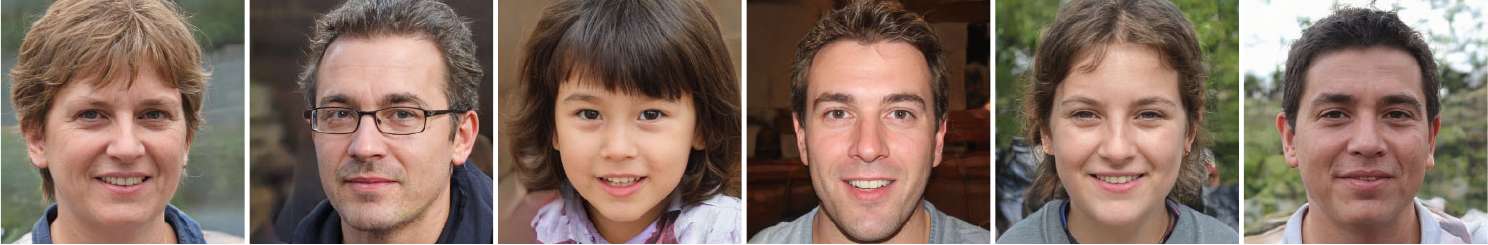}} \\
  \subfloat[VQGAN]{\includegraphics[width=\linewidth]{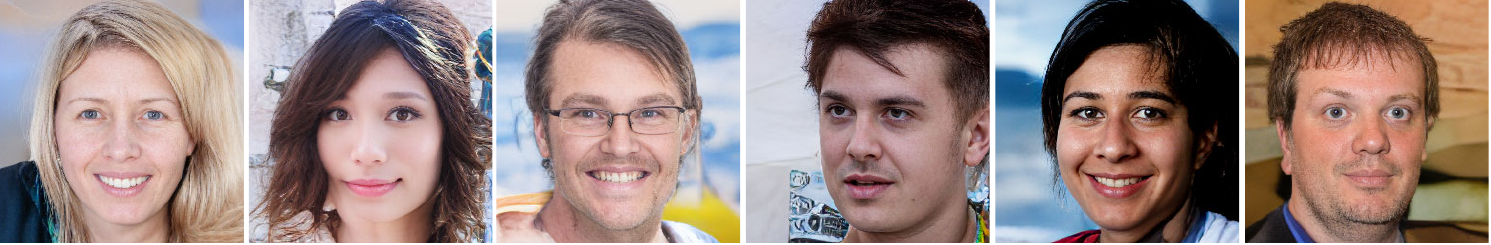}} \\
  \subfloat[LDM]{\includegraphics[width=\linewidth]{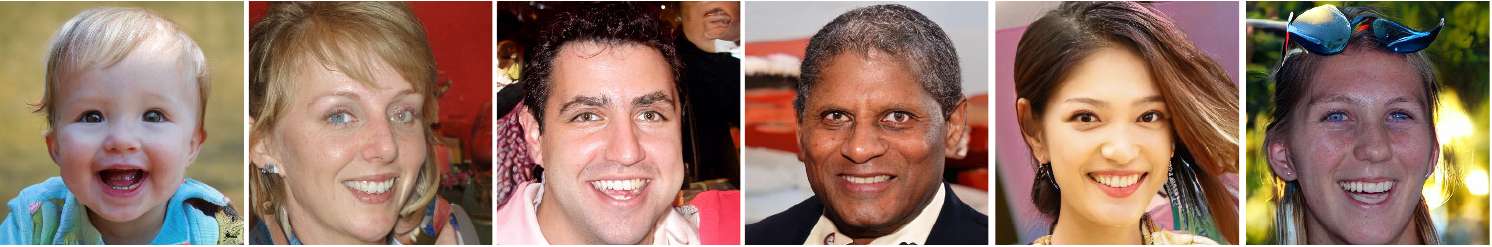}}
  \vspace{-0.2cm}
  \caption{Representative face images used in cross-dataset evaluation.}
  \label{supp_fig:face_cross_dataset}
\end{figure*}

\section{Competing Detectors}
\label{supp_sec: competing_methods}
This section briefly summarizes all competing detectors used for comparison. 

\noindent\textbf{CNND}~\cite{wang2019cnngenerated} trains a ResNet-50 classifier with standard data augmentations such as JPEG compression, Gaussian blurring as a way of improving generalizability.

\noindent\textbf{GramNet}~\cite{liu2020global} identifies AI-generated faces by capturing global texture statistics.

\noindent\textbf{RECCE}~\cite{Cao_2022_CVPR} learns to reconstruct face photographs. It is originally designed for face forgery detection, but is extended here to detect AI-generated faces.

\noindent\textbf{LNP}~\cite{liu2022detecting} extracts noise patterns using a pretrained denoising network, and fits a one-class support vector machine~\cite{scholkopf1999support_supp} to detect AI-generated faces as anomalies.

\noindent\textbf{LGrad}~\cite{tan2023learning} feeds gradient maps from a pretrained network as input to a detector to capture generative artifacts. 

\noindent\textbf{DIRE}~\cite{wang2023dire} assumes that diffusion models reconstruct synthetic images more accurately. Detection is based on reconstruction errors as input to a ResNet-50.

\noindent\textbf{Ojha23}~\cite{ojha2023towards} employs CLIP's frozen visual encoder to extract features for binary classification of photographic vs. AI-generated images.

\noindent\textbf{AEROBLADE}~\cite{ricker2024aeroblade} is a training-free approach that calculates LPIPS~\cite{zhang2018unreasonable_supp} reconstruction errors of latent diffusion autoencoders, leveraging the similar observation that synthetic images are reconstructed more faithfully.

\noindent\textbf{FatFormer}~\cite{liu2024forgeryaware} fine-tunes CLIP with a forgery-aware Transformer adapter, integrating spatial and frequency cues.

\noindent\textbf{Zou25}~\cite{zou2025self} casts ordinal EXIF-tag ranking as a pretext task, and employs one‑class anomaly detection for inference.

\noindent\textbf{CLIP}~\cite{radford2021CLIP} is a vision-language model pretrained on large-scale image-text pairs. Its general-purpose features can be adapted for identifying AI-generated content.

\noindent\textbf{FaRL}~\cite{zheng2022general} jointly learns signal-level and semantic-level face representations from photographic images via contrastive and masked modeling. Due to its transferable features, we tailor it for detecting AI-generated faces.

\noindent\textbf{EAL}~\cite{zheng2023exif} aligns EXIF metadata (as text prompts) with images, aiming to learn imaging-specific representations.

\noindent\textbf{Hu21}~\cite{hu2021exposing} detects GAN-generated faces by analyzing inconsistencies in corneal specular highlights, which are typically stable in human eyes but erratic in synthetic imagery.

\section{Visual Samples}
\label{supp_sec: dataset_vis}
To illustrate the diversity and practical relevance of our evaluation setups, we provide representative face images used during training and testing.

\begin{itemize}
    \item \textbf{Training Set:} Fig.~\ref{supp_fig:face_cross_model_1}(a) displays face photographs drawn from the CelebA-HQ dataset~\cite{karras2017progressive}. 
    \item \textbf{Cross-Generator Evaluation:} Figs.~\ref{supp_fig:face_cross_model_1}(b)-(j) show synthetic face images generated by nine representative models: StyleGAN2~\cite{karras2020analyzing}, VQGAN~\cite{esser2021taming}, LDM~\cite{rombach2022high}, DDIM~\cite{song2021denoising}, SDv2.1~\cite{rombach2022high}, FreeDoM~\cite{yu2023freedom}, HPS~\cite{wu2023better}, Midjourney~\cite{Midjourney}, and SDXL~\cite{podell2024sdxl}.
    \item \textbf{Cross-Dataset Evaluation:} Fig.~\ref{supp_fig:face_cross_dataset}  presents additional samples for domain transfer experiments. Specifically, the evaluation involves testing on FFHQ photographs and synthetic images generated by StyleGAN2, VQGAN, and LDM trained on FFHQ~\cite{karras2019style}. This setup assesses the robustness of the learned representations to variations in data distribution and image source.
\end{itemize}

These samples qualitatively demonstrate the visual similarity between photographic and AI-generated faces, highlighting the challenges of reliable detection and the necessity of learning discriminative, generalizable features.

\setcounter{enumi}{0}
\renewcommand\theenumi{S\arabic{enumi}}

\end{document}